\documentclass{IEEEtran}
\makeatletter
\let\MYcaption\@makecaption
\makeatother

\usepackage[font=footnotesize]{subcaption}

\makeatletter
\let\@makecaption\MYcaption
\makeatother
\usepackage{scalerel}
\usepackage{pgfplots}
\usepackage{filecontents}
\usepackage{pgfplotstable} 
\usepackage{tikz}
\usetikzlibrary{svg.path}
\usepackage{xcolor}
\definecolor{darkgreen}{rgb}{0.0, 0.5, 0.0} 
\usepgfplotslibrary{colorbrewer}
\usepgfplotslibrary{dateplot}
\pgfplotsset{compat=newest, width=8.5cm, height=6cm}
\usepackage{booktabs} 
\usepackage{siunitx} 
\usepackage{array}    
\usepackage{comment}
\usepackage{amsmath}
\usepackage{amsthm}
\usepackage{amssymb}
\usepackage{graphicx} 
\usetikzlibrary{arrows.meta, positioning, calc, bending, matrix} 
\usepackage[linesnumbered,ruled,vlined]{algorithm2e}

\definecolor{orcidlogocol}{HTML}{A6CE39}
\tikzset{
    orcidlogo/.pic={
        \fill[orcidlogocol] svg{M256,128c0,70.7-57.3,128-128,128C57.3,256,0,198.7,0,128C0,57.3,57.3,0,128,0C198.7,0,256,57.3,256,128z};
        \fill[white] svg{M86.3,186.2H70.9V79.1h15.4v48.4V186.2z}
        svg{M108.9,79.1h41.6c39.6,0,57,28.3,57,53.6c0,27.5-21.5,53.6-56.8,53.6h-41.8V79.1z M124.3,172.4h24.5c34.9,0,42.9-26.5,42.9-39.7c0-21.5-13.7-39.7-43.7-39.7h-23.7V172.4z}
        svg{M88.7,56.8c0,5.5-4.5,10.1-10.1,10.1c-5.6,0-10.1-4.6-10.1-10.1c0-5.6,4.5-10.1,10.1-10.1C84.2,46.7,88.7,51.3,88.7,56.8z};
    }
}

\newcommand\orcidicon[1]{\href{https://orcid.org/#1}{\mbox{\scalerel*{
                \begin{tikzpicture}[yscale=-1,transform shape]
                \pic{orcidlogo};
                \end{tikzpicture}
            }{|}}}}

\usepackage[colorlinks=true, linkcolor=blue, citecolor=blue, urlcolor=blue]{hyperref}

\usepgfplotslibrary{colormaps} 

\pgfplotsset{
    discard if not/.style 2 args={
        x filter/.expression={
            \thisrow{#1}==#2 ? x : nan
        }
    }
}

\newtheorem{proposition}{Proposition}[section]

\ifCLASSINFOpdf
\else
\fi
\begin{document}
%

\title{Advancing Explainable AI with Causal Analysis in Large-Scale Fuzzy Cognitive Maps}

%
%
%

\author{Marios~Tyrovolas$^{\textsuperscript{\orcidicon{0000-0002-4903-3204}}}$\,,~\IEEEmembership{Graduate Student Member,~IEEE,}
        Nikolaos~D. Kallimanis$^{\textsuperscript{\orcidicon{0000-0002-0331-1475}}}$,
        and~Chrysostomos~Stylios$^{\textsuperscript{\orcidicon{0000-0002-2888-6515}}}$\,,~\IEEEmembership{Senior~Member,~IEEE}
\thanks{This research has been financed by the European Union: Next Generation EU through the Program Greece 2.0 - National Recovery and Resilience Plan, under the call "Flagship actions in interdisciplinary scientific fields with a special focus on the productive fabric”, project name "Greece4.0 - Network of Excellence for developing, disseminating and implementing digital transformation technologies in Greek Industry" (project code: TAEDR-0535864).}
\thanks{Marios Tyrovolas, Nikolaos D. Kallimanis, and Chrysostomos Stylios are with the Department of Informatics and Telecommunications, University of Ioannina, Arta, 47 150 Greece, and the Industrial Systems Institute (ISI), Athena RC (e-mails: tirovolas@kic.uoi.gr, nkallima@isi.gr, stylios@isi.gr).}}

\maketitle

\begin{abstract}
In the quest for accurate and interpretable AI models, eXplainable AI (XAI) has become crucial. Fuzzy Cognitive Maps (FCMs) stand out as an advanced XAI method because of their ability to synergistically combine and exploit both expert knowledge and data-driven insights, providing transparency and intrinsic interpretability. This letter introduces and investigates the "Total Causal Effect Calculation for FCMs" (TCEC-FCM) algorithm, an innovative approach that, for the first time, enables the efficient calculation of total causal effects among concepts in large-scale FCMs by leveraging binary search and graph traversal techniques, thereby overcoming the challenge of exhaustive causal path exploration that hinder existing methods. We evaluate the proposed method across various synthetic FCMs that demonstrate TCEC-FCM's superior performance over exhaustive methods, marking a significant advancement in causal effect analysis within FCMs, thus broadening their usability for modern complex XAI applications.
\end{abstract}


%
\IEEEpeerreviewmaketitle

\section{Introduction}
%
%
%
%

\IEEEPARstart{R}{ecent} advancements in Artificial Intelligence (AI), Machine Learning (ML), and Deep Learning (DL) have revolutionized sectors such as healthcare, manufacturing, and finance, introducing algorithms with remarkable accuracy \cite{Jordan2015}. However, the "black-box" nature of many ML and DL models raises concerns about trust and transparency, particularly in sensitive applications, hindering their adoption \cite{Hou2022-nd}. This challenge has spurred the growth of eXplainable Artificial Intelligence (XAI), a field focused on making AI decisions transparent and understandable, thus increasing user trust, facilitating model refinement, and ensuring compliance with regulations \cite{Adadi2018-yp}. XAI encompasses both post-hoc explanation methods and intrinsic interpretable models. While post-hoc methods offer insights, they may misrepresent the underlying 'black-box' models' behavior \cite{Rudin2019-sg}, offer convincing explanations even for biased models \cite{Slack2020-ut}, or erroneously assume feature independence \cite{Aas2021-mq}. These issues underscore the importance of developing intrinsically interpretable models that incorporate causality, offering explanations that reflect authentic feature interactions, minimize bias, and resonate with human intuition, further advancing XAI \cite{Liang2023-bj}.

Among the interpretable techniques, Fuzzy Cognitive Maps (FCMs) stand out for their ability to model complex systems, perform what-if simulations, and conduct predictive tasks \cite{Kosko1986-zl}. FCMs can be expressed as recurrent neural networks structured as directed graphs with nodes called concepts that represent the main components of the modelled system and weighted edges that describe the causal relations between concepts \cite{Kosko1986-zl}, thus offering an intuitive representation of the system under analysis. FCMs, developed based on expert knowledge \cite{Stylios2004-us} and/or learning algorithms \cite{Tyrovolas2023-dt}, excel in scenarios with limited data availability due to their unique ability to combine expert and data-driven insights \cite{Kreinovich2015-kt}. Enabled by fuzzy logic, this synergy also grants experts the flexibility to adjust model weights, thereby incorporating new insights not previously reflected in the data \cite{Napoles2023-ea}. Notably, FCMs provide a transparent inference process and inherently offer feature-based explanations for their decisions from global and local perspectives \cite{Napoles2023-jf}.

Global explainability in FCMs is achieved through static analysis, which delves into the model's structure, concentrating on the interconnection weights among concepts to reveal the system's complexity and highlight key concepts and their causal influences. For this purpose, Kosko \cite{Kosko1986-zl} introduced a simple fuzzy causal algebra designed for calculating the indirect and total causal effects among concepts. Specifically, this methodology includes identifying all the directed causal paths linking the concepts, assessing the indirect effect along each path, and subsequently using these effects to identify the total effect of the causative concept.

Traditionally, this in-depth analysis within an FCM's directed graph, represented as $\mathcal{G} = (C, E)$ with $C$ and $E$ signifying the set of concepts and edges respectively, and \(n = |C|\), \(e = |E|\) denote their counts, requires exhaustive search algorithms to identify all causal paths between any two nodes  \cite{freund2021necessity}. However, in cases where FCMs are fully interconnected, this method necessitates examining every possible order in which concepts can be visited, each representing a unique path. As a result, the computational complexity can soar to $\mathcal{O}(n!)$ \cite{Sryheni_2022}, making static analysis infeasible for even relatively small-scale FCMs and confining its utility \cite{Dodurka2017-rk}. This limitation highlights the need for more efficient approaches to analyzing the causal effects within FCMs, including those of large scale, ensuring their practicality across a wider range of applications. To the best of our knowledge, no such approach has been proposed until now, highlighting a significant gap in the field.

To address this computational bottleneck, this letter introduces the \textit{"Total Causal Effect Calculation for FCMs"} (TCEC-FCM) algorithm, an innovative methodology that provides efficient calculation of the total causal effect in large-scale FCMs without requiring exhaustive path exploration, and operates with a computational complexity of $\mathcal{O}(n \cdot e \cdot \log e)$. Uniquely, the TCEC-FCM algorithm is the first to address this challenge, setting a precedent in the field. Its effectiveness was validated experimentally on large-scale synthetic FCMs of various sizes and densities, demonstrating substantial computational savings compared to traditional exhaustive methods. This pioneering contribution not only addresses a critical gap in FCM analysis but also broadens the practical applicability of FCMs in XAI.

\section{Preliminaries}

\subsection{Fuzzy Cognitive Maps}
\begin{figure}
\centering
\begin{tikzpicture}[>=Stealth, node distance=2cm, auto, scale=0.81, every node/.style={scale=0.81}]
\node (C1) [draw, circle] {$C_{1}$};
\node (C2) [draw, circle, below left of=C1] {$C_{2}$};
\node (C3) [draw, circle, above right of=C1] {$C_{3}$};
\node (C4) [draw, circle, below right of=C1] {$C_{4}$};

\draw[->] (C1) -- node[scale=0.75] {+0.6} (C3);
\draw[->] (C2) -- node[scale=0.75] {+0.68} (C1);
\draw[->] (C2) -- node[scale=0.75, swap,above] {-0.7} (C4);
\draw[->] (C4) -- node[scale=0.75, swap] {+0.36} (C3);
\draw[->] (C3) to[bend left] node[scale=0.75, pos=0.5, sloped, below] {+0.15} (C1);
\draw[->] (C4) to[bend left] node[scale=0.75, swap,below] {-0.25} (C2); 

\node (matrix) [right=1cm of C3, anchor=west, yshift=-1.5cm] {
$\mathbf{W} =
\begin{bmatrix}
0 & 0 & +0.6 & 0 \\
+0.68 & 0 & 0 & -0.7 \\
+0.15 & 0 & 0 & 0 \\
0 & -0.25 & +0.36 & 0
\end{bmatrix}$
};

\end{tikzpicture}
\caption{Example FCM with four concepts.}
\label{fig:fcm_example}
\end{figure}
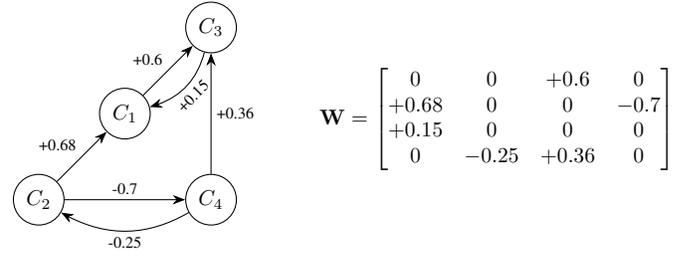

This section outlines the fundamentals of FCMs, focusing on their structure and dynamics. FCMs consist of $n$ concepts $C_{i} \:, i \in \{1,2,\dotso,n\}$ and weighted connections, $w_{ij} \in [-1,1]$, which indicate the causal impact of $C_{i}$ on $C_{j}$ (Fig.~\ref{fig:fcm_example}). The causal relationships are categorized as \textbf{positive} ($w_{ij}>0)$, where $C_{j}$ changes in the same direction as $C_{i}$, \textbf{negative} ($w_{ij}<0)$, where $C_{j}$ changes in the opposite direction, or \textbf{zero} ($w_{ij}=0)$, indicating no relation. Each concept $C_{i}$ has an activation value $A_{i}$ within $[0,1]$ or $[-1,1]$, calculated  using a reasoning rule, with the most prevalent being:

\begin{equation}
\label{reasoning rule}
A_{i}^{(t+1)}=f(\sum_{\substack{j =1 \\ j \neq i}}^n A_{j}^{(t)} w_{ji}),
\end{equation}

In \eqref{reasoning rule}, $t$ represents the iteration step, ranging from 0 to a user-defined maximum $T$. $A_{i}^{(t+1)}$ and $A_{j}^{(t)}$ are activation values of concepts $i$ and $j$ at steps $t+1$ and $t$ respectively, $w_{ji}$ the weight from concept $j$ to $i$, and $f(\cdot)$ the activation function normalizing concept values to the allowed activation interval. Common activation functions include bivalent, trivalent, hyperbolic tangent, and sigmoid.

At each iteration step $t$, the concepts' activation values form a state vector $\mathbf{A}^{(t)} \in \mathbb{R}^{n}$, and the causal weights $w_{ij}$ compose a weight matrix $\mathbf{W} \in \mathbb{R}^{n \times n}$ with zero diagonal elements. Therefore, \eqref{reasoning rule} can be rewritten as:
\begin{equation}
\mathbf{A}^{(t)}=f(\mathbf{A}^{(t-1)} \mathbf{W}).
\label{eq_3}
\end{equation}

 The FCM's iterative reasoning process begins with an initial state vector $\mathbf{A}^{(0)}$, provided by domain experts or extracted from a dataset, and the employed reasoning rule is then applied recurrently until the FCM converges to a fixed point attractor (by $t \leq T$) or reaches the maximum number of iterations ($T$), indicating cyclic or chaotic behavior.

\subsection{Indirect and Total Causal Effect Analysis}

In this subsection, we introduce the concept of indirect and total causal effects between concepts, following Kosko's fuzzy causal algebra \cite{Kosko1986-zl}. This mathematical formulation is vital for understanding the complex interactions within FCMs, laying the groundwork for our subsequent analysis and algorithmic development.

\begin{proposition}[\cite{Kosko1986-zl}]
Consider $m$ causal paths from concept $C_{i}$ to $C_{j}$, each denoted as ($C_i \rightarrow C_{k_{1}^{l}} \rightarrow \ldots \rightarrow C_{k_{n}^{l}} \rightarrow C_j$), where $1 \leq l \leq m$ and \( C_{k_{x}^{l}} \) are intermediate concepts within the \( l^{th} \) causal path. The indirect effect of $C_{i}$ on $C_{j}$ along a specific path $l$ is equal to the weakest link within that sequence. This is mathematically represented as:

\begin{equation}
    I_{l}(C_{i},C_{j})=min\{w(C_{p},C_{p+1})\},
    \label{indirect_effect}
\end{equation}
where $w(C_p, C_{p+1})$ indicates the causal weight between each pair of consecutive concepts \( C_p \) and \( C_{p+1} \) along the sequence. Finally, the total causal effect $T(C_{i},C_{j})$ is the maximum of these indirect effects:

\begin{equation}
    T(C_{i},C_{j})=max\{I_{l}(C_{i},C_{j})\}, 1 \leq l \leq m
    \label{total_effect}
\end{equation}
\label{proposition_2_1}
\end{proposition}

Drawing from theoretical insights, our objective is to devise an algorithm that enables the efficient calculation of total causal effects within large-scale FCMs, thereby overcoming the exponential complexity associated with exhaustive search techniques.

\section{Total Causal Effect Calculation for FCMs (Tcec-Fcm)}

\subsection{Conceptual Framework and Strategy} \label{conceptual_framework}

This section presents the TCEC-FCM algorithm designed for calculating the total causal effect, $T(C_{i}, C_{j})$, between two concepts, $C_{i}$ and $C_{j}$ in large-scale FCMs. At the core of TCEC-FCM's methodology is the initial step of sorting the weights in descending order. Subsequently, for each concept $C_{i}$, a copy of the FCM is initialized, where all concepts are isolated. Weights from the sorted list are then incrementally added to this copy, starting with the highest value. After each addition, graph traversal ascertains whether a causal path between $C_{i}$ and $C_{j}$ has been established. The first weight that enables such a connection is considered critical, serving as the minimal weight in the identified path, because all prior weights are larger; thus, it denotes the indirect effect $I_{l}(C_{i}, C_{j})$. In addition, because this weight is greater than all subsequent weights, it also denotes the maximum possible indirect effect $T(C_{i}, C_{j})$. Therefore, TCEC-FCM's primary aim is to identify this initial, path-establishing weight in the descending sequence.

To implement this concept, we incorporate a hybrid approach, combining binary search for locating the critical weight and Breadth-First Search (BFS) for comprehensive exploration of potential causal paths. Binary search, with its $\mathcal{O}(\log e)$ time complexity, enhances efficiency compared to methods like linear search, which has $\mathcal{O}(e)$ complexity. This efficiency, combined with its deterministic, space-efficient, and simplistic nature, makes it well-suited for the developed algorithm. On the other hand, BFS is selected for its systematic exploration of nodes at each depth level, providing thorough coverage within the FCM, rather than for any clear superiority over alternatives such as Depth-First Search (DFS) or Linear-First Search (L-FS).

\subsection{Practical Implementation and Algorithmic Workflow}

\begin{figure*}
  \centering
  \includegraphics[width=\textwidth]{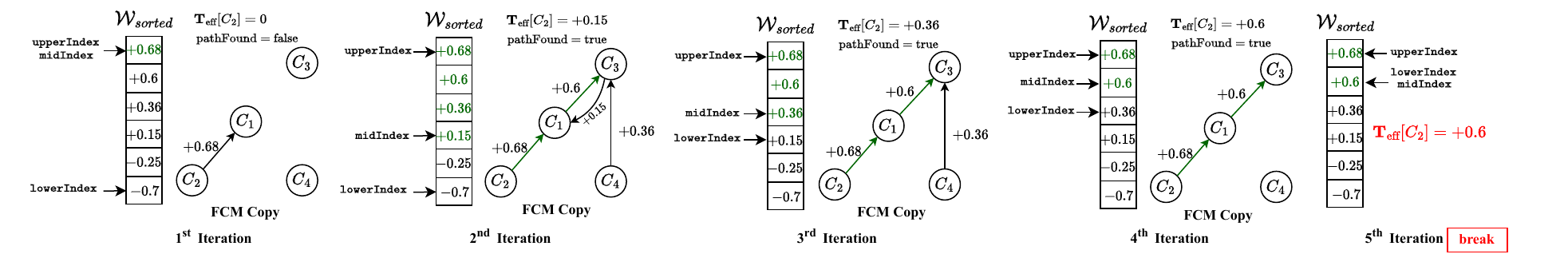}
  \caption{Step-by-step illustration of TCEC-FCM for the example of Fig.\ref{fig:fcm_example}.}
  \label{fig:algorithm}
\end{figure*}

The practical implementation of the TCEC-FCM algorithm, detailed in Algorithm~\ref{tcec_fcm_pseudocode}, begins by organizing the FCM weights into a descending column vector, $\mathcal{W}_{sorted}$, and applies binary search for each concept $C_{i}$, using the indices \texttt{upperIndex}, \texttt{midIndex}, and \texttt{lowerIndex} to manage the search space. Initially, \texttt{upperIndex} and \texttt{midIndex} are positioned at the first element of $\mathcal{W}_{sorted}$, and \texttt{lowerIndex} at the last (lines 5-7). In the iterative phase, weights from $\mathcal{W}_{sorted}$ are dynamically added to an FCM copy until the weight indicated by \texttt{midIndex} is reached (lines 10-13). Subsequently, BFS checks for a causal path between $C_{i}$ and $C_{j}$ in this dynamic FCM copy (line 14). If no path is found, the algorithm narrows the search between \texttt{midIndex} and \texttt{lowerIndex} (line 20); if found, the focus shifts to the upper half, marking the \texttt{midIndex} weight as temporary $T(C_{i}, C_{j})$ and setting a '$\text{pathFound}$' flag (lines 16-18). Once the search interval has been adjusted, \texttt{midIndex} is set as the new interval's midpoint (line 21). This process is repeated until the search interval cannot be reduced further, at which point the weight at \texttt{midIndex} is finalized as $T(C_{i}, C_{j})$. If the algorithm completes without identifying a path, $T(C_{i}, C_{j})$ is set to zero (lines 24-25). Fig.~\ref{fig:algorithm} visually outlines this process, as applied to the FCM in Fig.~\ref{fig:fcm_example}.

\subsection{Time and Space Complexity Analysis}


The TCEC-FCM algorithm's time complexity arises from its preprocessing and core computational steps. Initially, preprocessing identifies \(e\) non-zero weights from $W$'s \(n^2\) elements in \(\mathcal{O}(n^2)\) time and sorts them in \(\mathcal{O}(e \cdot \log e)\) using time-optimal sorting algorithm like \textit{MergeSort} (line 2). Secondly, the main for-loop (line 4) is executed $n-1$ times, while the nested while-loop (line 8) for binary search on $\mathcal{W}_{sorted}$ operates within $\mathcal{O}(\log e)$ iterations. Moreover, the execution cost of BFS, which is typically $\mathcal{O}(n + e)$ in arbitrary graphs, simplifies to $\mathcal{O}(e)$ in cases where $n < e$. Additionally, the worst-case scenario for the for-loop of line 11 involves $\mathcal{O}(e)$ iterations, each with constant operational steps. Thus, the for-loop of line 4 contributes $\mathcal{O}(n \cdot e \cdot \log e)$ to the total time complexity. Consequently, the overall time complexity is  $\mathcal{O}(n^2 + e \cdot \log e + n \cdot e \cdot \log e) \in \mathcal{O}(n \cdot e \cdot \log e)$. Regarding space complexity, TCEC-FCM employs two 2D arrays ($\mathbf{W}$ and $\mathbf{W}_{\text{copy}}$), and two 1D arrays ($\mathcal{W}_{nz}$ and $\mathcal{W}_{sorted}$), each with a maximum capacity of $n^2$ elements, alongside a constant number of variables. This results in a space complexity of $\mathcal{O}(n^2)$.

\begin{small} 
\begin{algorithm}[t]
\DontPrintSemicolon
\caption{Total Causal Effect Calculation for Fuzzy Cognitive Maps (TCEC-FCM)}
\label{tcec_fcm_pseudocode}

\KwIn{
    $n$, number of concepts; $\mathbf{W_{n \times n}}$, weight matrix.
}
\KwOut{
    $\mathbf{T}_{\text{eff}}$, vector of total causal effects.
}
\tcp{Extract and sort non-zero weights and their indices}
\( \mathcal{W}_{nz} \gets \{(i,j,w_{ij}) \mid w_{ij} \in W, w_{ij} \neq 0\} \)\;
$\mathcal{W}_{sorted} \gets \text{sortDescending}(\{w_{ij} \mid (i,j,w_{ij}) \in \mathcal{W}_{nz}\})$ \CommentSty{// $\mathcal{W}_{sorted} \in \mathbb{R}^{e \times 1}$ where $e$ is the number of non-zero weights}

\tcp{Initialize $\mathbf{T}_{\text{eff}}$}
$\mathbf{T}_{\text{eff}} \gets [0]_{n \times 1}$\;

\For{$C_i \leftarrow 1$ \KwTo $n-1$}{
    \tcc{Init binary search vars}
    $\text{exIdxs} \gets \emptyset, \text{pathFound} \gets \text{false}$\;
    $\text{upperIndex} \gets 1, \text{midIndex} \gets 1$\;
    $\text{lowerIndex} \gets \text{length}(\mathcal{W}_{\text{sorted}})$\;
    \While{$upperIndex - lowerIndex \geq 1$}{
        $\text{exIdxs} \gets \text{exIdxs} \cup \{\text{midIndex}\}$\;
        \tcc{FCM copy for weights subset}
        $\mathbf{W}_{\text{copy}} \gets [0]_{n \times n}$ \;

        \For{$idx \gets 1$ \KwTo $\text{midIndex}$}{
            $(i,j,w_{ij}) \gets \mathcal{W}_{\text{sorted}}[idx]$\;
            $\mathbf{W}_{\text{copy}}[i][j] \gets w_{ij}$\;
        }
        \tcc{BFS for reachability to $C_n$}
        $\text{reachableConcepts} \gets \text{BFS}(\mathbf{W}_{\text{copy}}, C_i, C_n)$\;
        \eIf{$C_{n} \in \text{reachableConcepts}$}{
            $\mathbf{T}_{\text{eff}}[C_i] \gets \mathcal{W}_{\text{sorted}}[\text{midIndex}]$\;
            $\text{lowerIndex} \gets \text{midIndex}$\;
            $\text{pathFound} \gets \text{true}$\; 
        }{
            $\text{upperIndex} \gets \text{midIndex}$\;
        }
        $\text{midIndex} \gets \text{Round}\left(\frac{\text{upperIndex} + \text{lowerIndex}}{2}\right)$\;
        \tcc{Check for convergence}
        \If{$(\text{upperIndex} - \text{lowerIndex}) = 1$ \textbf{and} $midIndex \in \text{exIdxs}$}{
            \textbf{break}\; 
        }
    }
    \If{\textbf{not} \text{pathFound}}{
        $\mathbf{T}_{\text{eff}}[C_i] \gets 0$ \CommentSty{// No path found}\; 
        
    }

}
\Return $\mathbf{T}_{\text{eff}}$\;
\end{algorithm}
\end{small}

\section{Experimental Results}

This section evaluates the performance of the proposed TCEC-FCM algorithm by comparing its execution times on synthetic FCMs of various sizes and densities against the TCEC-FCM-LS algorithm, its linear search-based variant, and DFSB-FCM, an exhaustive algorithm that combines DFS and backtracking to dynamically update $T(C_{i},C_{j})$ with the highest $I_{l}(C_{i},C_{j})$ discovered through causal path analysis. Initial evaluations focused on fully interconnected FCMs (100\% density) to simulate the worst-case scenario (WCS). The study was then broadened to encompass FCMs of varied densities. All experiments were carried out on a desktop computer equipped with a 12-core AMD Ryzen 9 5900X processor and 128GB of RAM, using MATLAB for computational tasks. The source code for these experiments can be found at \cite{tyrovolas_2024_10613580}.

\subsection{Execution Time Analysis in WCS}

In the WCS analysis, TCEC-FCM-LS and TCEC-FCM were evaluated on FCMs with 10, 100, and 1000 nodes, while DFSB-FCM was confined to 10-, 12-, and 13-node FCMs due to its exhaustive search nature. To ensure comparability, 40 random FCMs were generated for the various node sizes for the three algorithms. The average execution times, measured using MATLAB's tic-toc function, are summarized in Tab.~\ref{tab:execution_times_wcs}. The results clearly show that both TCEC-FCM algorithms significantly outperform DFSB-FCM across all sizes. In detail, TCEC-FCM-LS and TCEC-FCM maintained execution times below one second for up to 100 nodes, while DFSB-FCM's times escalated to 6822.280 seconds for just 13 nodes.  For 1000-node FCMs, TCEC-FCM-LS and TCEC-FCM completed in under 635 seconds, highlighting their efficiency in large-scale networks, in contrast to DFSB-FCM's prohibitively high execution times, denoted by {\small $\ggg$}.

The experiments also revealed that TCEC-FCM-LS's performance is significantly influenced by the distribution of weight values in $\mathcal{W}_{sorted}$. In fact, it operates efficiently when solutions are located early in the sorted list, but slows down for lower-ranked solutions due to the sequential search mechanism. On the other hand, TCEC-FCM employs a binary search strategy that consistently reduces its search space by half, minimizing the effect of weight distribution and thus maintaining more uniform execution times. This suggests TCEC-FCM's reliability and effectiveness for large-scale FCMs, contrasting with TCEC-FCM-LS's unpredictable performance, as depicted in Fig.~\ref{fig:execution_times_variability}.

\begin{table}
\centering
\caption{Average Execution Times (s) per Concept Count for the Worst-Case Scenario}
\label{tab:execution_times_wcs}
\setlength{\tabcolsep}{7.5pt} 
\begin{tabular}{@{}l@{\hspace{7.5pt}}ccccc@{}}
\toprule
\textbf{Algorithm} & \multicolumn{5}{c}{\textbf{FCM concepts}} \\
\cmidrule{2-6}
                    & \textbf{10} & \textbf{12} & \textbf{13} & \textbf{100} & \textbf{1000} \\
\midrule
TCEC-FCM-LS        & \textbf{0.025}    & \textbf{0.025}
    & \textbf{0.025}
    & \textbf{0.697} & \textbf{558.386} \\
TCEC-FCM        & 0.027    & 0.030
    & 0.032
    & 1.307 & 634.807 \\
DFSB-FCM           & 5.820  & 588.410  & 6822.280  & $\ggg$    & $\ggg$ \\
\bottomrule
\end{tabular}
\raggedright
\footnotesize{Note: $\ggg$ indicates excessive computation time.}
\end{table}

\begin{figure}
\centering
\begin{tikzpicture}
\begin{axis}[
    xlabel={\small Trial Number},
    ylabel={\small Execution Time (s)},
    legend pos=north east,
    ymajorgrids=true,
    grid style=dashed,
    width=0.6\columnwidth,
    height=3.5cm,
    scale only axis,
    legend style={font=\small},
    tick label style={font=\scriptsize}, 
    cycle list name=black white, 
]

\addplot+[mark=square*, every mark/.append style={fill=gray}] table[
    col sep=comma,
    x=Trial,
    y=TCEC-FCM-BS,
] {variability_data.csv};
\addlegendentry{\small TCEC-FCM}

\addplot+[mark=*, every mark/.append style={fill=black}] table[
    col sep=comma,
    x=Trial,
    y=TCEC-FCM-LS,
] {variability_data.csv};
\addlegendentry{\small TCEC-FCM-LS}

\end{axis}
\end{tikzpicture}
\caption{Execution times of TCEC-FCM across fully interconnected FCM trials with 1000 concepts.}
\label{fig:execution_times_variability}
\end{figure}
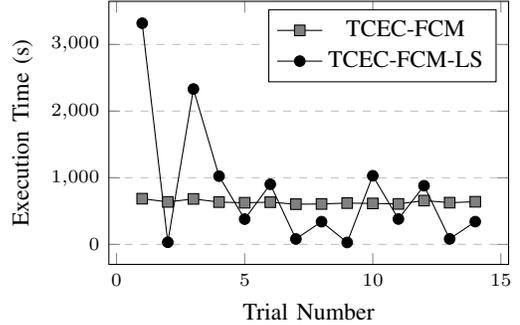

\begin{figure*}
\begin{minipage}[b]{0.32\linewidth}
\centering
\begin{tikzpicture}[scale=0.68]
\begin{axis}[
    xlabel={Density},
    ylabel={Average Execution Time (s)},
    ymin=0, ymax=7500, 
    ymajorgrids=true,
    extra y ticks={524},
    extra y tick style={
        grid style={darkgreen, dashed},
        tick label style={color=darkgreen, anchor=east}
    },
    grid style=dashed,
    legend pos=north west,
    legend cell align={left},
    legend style={fill=none, font=\small}, 
    cycle list name=color list
]

\addplot+[
    discard if not={FCM_Concepts}{10},
    blue,
    mark=*
] table [
    col sep=comma,
    x=Density,
    y=AVERAGE
] {3D_TCEC_FCM_LS_FINAL_RESULTS.csv};
\addlegendentry{$n=10$}

\addplot+[
    discard if not={FCM_Concepts}{100},
    red,
    mark=square*
] table [
    col sep=comma,
    x=Density,
    y=AVERAGE
] {3D_TCEC_FCM_LS_FINAL_RESULTS.csv};
\addlegendentry{$n=100$}

\addplot+[
    discard if not={FCM_Concepts}{1000},
    black,
    mark=triangle*,
] table [
    col sep=comma,
    x=Density,
    y=AVERAGE
] {3D_TCEC_FCM_LS_FINAL_RESULTS.csv};
\addlegendentry{$n=1000$}

\end{axis}
\end{tikzpicture}
\subcaption{TCEC-FCM-LS}\label{fig:TCEC_FCM_LS}%
\end{minipage}%
\hfil
\begin{minipage}[b]{0.32\textwidth}
\centering
\begin{tikzpicture}[scale=0.68]
\begin{axis}[
    xlabel={Density},
    ylabel={Average Execution Time (s)},
    ymin=0, ymax=7500, 
    ymajorgrids=true,
    extra y ticks={524},
    extra y tick style={
        grid style={darkgreen, dashed},
        tick label style={color=darkgreen, anchor=east}
    },
    grid style=dashed,
    legend pos=north west,
    legend cell align={left},
    legend style={fill=none, font=\small}, 
    cycle list name=color list,
]

\addplot+[
    discard if not={FCM_Concepts}{10},
    blue,
    mark=*
] table [
    col sep=comma,
    x=Density,
    y=AVERAGE
] {3D_TCEC_FCM_BS_FINAL_RESULTS.csv};
\addlegendentry{$n=10$}

\addplot+[
    discard if not={FCM_Concepts}{100},
    red,
    mark=square*
] table [
    col sep=comma,
    x=Density,
    y=AVERAGE
] {3D_TCEC_FCM_BS_FINAL_RESULTS.csv};
\addlegendentry{$n=100$}

\addplot+[
    discard if not={FCM_Concepts}{1000},
    black,
    mark=triangle*
] table [
    col sep=comma,
    x=Density,
    y=AVERAGE
] {3D_TCEC_FCM_BS_FINAL_RESULTS.csv};
\addlegendentry{$n=1000$}

\end{axis}
\end{tikzpicture}
\subcaption{TCEC-FCM}\label{fig:TCEC_FCM_BS}%
\end{minipage}%
\hfil
\begin{minipage}[b]{0.32\textwidth}
\centering
\begin{tikzpicture}[scale=0.68]
\begin{axis}[
    xlabel={Density},
    ylabel={Average Execution Time (s)},
    ymin=0, ymax=7500, 
    ymajorgrids=true,
    grid style=dashed,
    legend pos=north west,
    legend cell align={left},
    legend style={fill=none, font=\small}, 
    cycle list name=color list,
]

\addplot+[
    discard if not={FCM_Concepts}{10},
    blue,
    mark=*
] table [
    col sep=comma,
    x=Density,
    y=AVERAGE
] {3D_EXH_FINAL_RESULTS.csv};
\addlegendentry{$n=10$}

\addplot+[
    discard if not={FCM_Concepts}{12},
    red,
    mark=square*
] table [
    col sep=comma,
    x=Density,
    y=AVERAGE
] {3D_EXH_FINAL_RESULTS.csv};
\addlegendentry{$n=12$}

\addplot+[
    discard if not={FCM_Concepts}{13},
    black,
    mark=triangle*
] table [
    col sep=comma,
    x=Density,
    y=AVERAGE
] {3D_EXH_FINAL_RESULTS.csv};
\addlegendentry{$n=13$}

\end{axis}
\end{tikzpicture}
\subcaption{DFSB-FCM}\label{fig:DFSB_FCM}%
\end{minipage}%
\caption{Average execution times of compared algorithms at varying FCM densities.}
\label{fig:execution_time_trends}
\end{figure*}
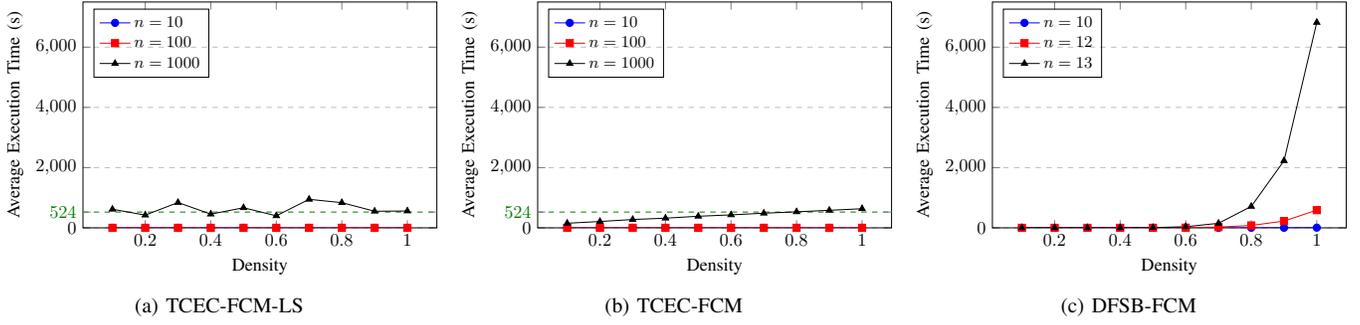

\subsection{Execution Time Analysis Across Various FCM Densities}

In the next phase, we extended our analysis to investigate simulations on FCMs with densities varying from 10\% to 100\% and with the same node counts: 10, 100, and 1000 nodes for both TCEC-FCM-LS and TCEC-FCM, and 10, 12, and 13 nodes for DFSB-FCM. Consistent with the previous analysis, 40 random FCMs were created for each combination of node count and density. Initially, the performance assessment across these densities required the calculation of the mean execution time for each algorithm at the highest node count tested. This involved firstly determining the average execution time at each density level, which was then consolidated to yield a singular mean execution time value per algorithm using the formula:

\begin{equation}
    \bar{T}_{\text{algorithm}, n} = \frac{1}{D} \sum_{d=1}^{D} \bar{T}_{\text{algorithm}, n, d},
\end{equation}
where \( \bar{T}_{\text{algorithm}, n} \) represents the overall average execution time for an algorithm at node count \( n \), \( D \) is the total number of densities analyzed, and \( \bar{T}_{\text{algorithm}, n, d} \) is the average execution time at node count \( n \) and density \( d \). The findings presented in Tab.~\ref{tab:algorithm_mean_time} highlight the superior performance and suitability of the TCEC-FCM algorithm for large-scale FCM applications. TCEC-FCM-LS still achieves commendable execution times for a large number of concepts while exhibiting less efficiency owing to its variability in behavior. In contrast, DFSB-FCM, with its markedly higher average execution times even at small node counts, is not applicable in large-scale scenarios. This comparison underscores TCEC-FCM's robustness and effectiveness for complex and large FCM analyses, positioning it as the preferred choice.

\begin{table}
\centering
\caption{Mean Execution Time per Algorithm Across Densities}
\label{tab:algorithm_mean_time}
\begin{tabular}{
  l
  S[table-format=3.2]
  S[table-format=3.2]
  S[table-format=3.2]
} 
\toprule
{Algorithm} & {TCEC-FCM-LS} & {TCEC-FCM} & {DFSB-FCM} \\
\toprule
{Node count} & {\( 1000 \)} & {\( 1000 \)} & {\( 13 \)} \\
{Execution time (s)} & 629.55 & {\bfseries 400.95} & 995.93 \\
\bottomrule
\end{tabular}
\end{table}

Further analysis, as depicted in Fig.~\ref{fig:execution_time_trends}, shows the average execution time of each algorithm against FCM density. The TCEC-FCM algorithms consistently maintain low execution times across different densities, highlighting their efficiency. In contrast, DFSB-FCM's execution time spikes sharply for densities above 80\%, suggesting its limitation for sparse, smaller-scale FCMs. As in WCS, TCEC-FCM-LS's performance varies with the weight distribution in \(\mathcal{W}_{sorted}\), in contrast to TCEC-FCM's uniform execution times, which only show minor increases in execution time as the node count and density increase. This distinction is further clarified by a benchmark green dashed line indicating an average median execution time of the TCEC-FCM algorithms of approximately 523.87 seconds. As can be observed, the execution time of TCEC-FCM-LS fluctuates around this benchmark, underlining its inconsistency, while TCEC-FCM remains consistently below this mark, affirming its reliable performance even as the density increases.

\section{Conclusions}
This letter introduces the TCEC-FCM algorithm, a novel method for conducting causal effect analysis within large-scale FCMs, to enhance and broaden their usability in XAI applications with numerous variables. Our detailed evaluation across synthetic FCMs of varying sizes and densities showcases the algorithm's superior performance over traditional exhaustive methods in every tested scenario. However, by extending our experiments to very-large-scale FCMs exceeding 10,000 concepts, the reasonable yet extended execution times achieved for such vast models highlight a pivotal area for future research. In future, we plan to leverage parallel computing and high-performance programming languages, such as C, to refine the execution speeds for these massive models, aiming to broaden the usability and scalability of FCMs within the XAI domain.


%

\ifCLASSOPTIONcaptionsoff
  \newpage
\fi



\bibliographystyle{IEEEtran}
\bibliography{letterbib}

%




\end{document}